\documentclass[numbers]{article}

\usepackage[preprint]{main}
\usepackage[utf8]{inputenc} 
\usepackage[T1]{fontenc}    
\usepackage{hyperref}       
\usepackage{url}            
\usepackage{booktabs}       
\usepackage{amsfonts}       
\usepackage{amsmath}       
\usepackage{nicefrac}       
\usepackage{microtype}      
\usepackage{xcolor}         
\usepackage{graphicx}
\usepackage{subcaption} 

\title{ScriptViT: Vision Transformer-Based Personalized Handwriting Generation}

\author{%
  Sajjan Acharya\thanks{Equal contribution.} \\
  Independent Researcher \\
  \texttt{sajjanacharya00@gmail.com}
  \And
  Rajendra Baskota\footnotemark[1] \\
  Independent Researcher \\
  \texttt{raj.baskota056@gmail.com}
}

\begin{document}
\maketitle


\begin{abstract}
  Styled handwriting generation aims to synthesize handwritten text that looks both realistic and aligned with a specific writer’s style. While recent approaches involving GAN, transformer and diffusion-based models have made progress, they often struggle to capture the full spectrum of writer-specific attributes, particularly global stylistic patterns that span long-range spatial dependencies. As a result, capturing subtle writer-specific traits such as consistent slant, curvature or stroke pressure, while keeping the generated text accurate is still an open problem. In this work, we present a unified framework designed to address these limitations. We introduce a Vision Transformer-based style encoder that learns global stylistic patterns from multiple reference images, allowing the model to better represent long-range structural characteristics of handwriting. We then integrate these style cues with the target text using a cross-attention mechanism, enabling the system to produce handwritten images that more faithfully reflect the intended style. To make the process more interpretable, we utilize Salient Stroke Attention Analysis (SSAA), which reveals the stroke-level features the model focuses on during style transfer. Together, these components lead to handwriting synthesis that is not only more stylistically coherent, but also easier to understand and analyze.
\end{abstract}

\section{Introduction}
Handwriting generation based on typed text and conditioned on a writer's calligraphic style is a challenging problem. Automatic generation of handwritten words in a person's original style has applications in generating synthetic data for building signature verification and handwritten text recognition systems. It can also be used for the restoration of degraded or incomplete documents by filling in the missing regions while preserving their original calligraphic characteristics. However, achieving realistic handwriting that preserves textual accuracy and a writer's local and global style remains a challenging task. This is due to the fact that a writer's style comprises many subtle elements that must be captured for effective imitation. Characteristics such as consistent slant, stroke thickness and pressure, curvature and character morphology all contribute to the distinctiveness of personal handwriting. Handwriting exhibits both fine-grained local properties and broader global structures, with long-range relationships between strokes playing a significant role in defining a writer's overall style. Capturing these multi-scale attributes requires a model that can interpret local stroke details while maintaining a coherent understanding of the global spatial arrangement of characters and words.

Previous frameworks for offline handwriting generation typically utilize CNN-based GAN architectures \cite{ref3, ref5, ref13, ref14, ref15} to model style-content entanglement, but they often struggle to establish a strong and coherent connection between the provided style and the target text. To capture this entanglement at the character level, Bhunia et al. \cite{ref2} introduce a transformer-based architecture for styled handwriting generation, where they utilize a hybrid CNN and transformer-based encoder to capture the style components. Pippi et al. \cite{ref17} and Vanherle et al. \cite{ref21} further extend this work by modifying the context text representation and studying the impact of both visual and textual components on the HTG model training. Following the success of CNN and transformer-based Handwriting Text Generation (HTG), Nikolaidou et al. \cite{ref16} and Dai et al. \cite{ref19} utilize diffusion models to effectively imitate the writer’s style. However, diffusion-based HTG models require many iterative denoising steps to generate a single image, resulting in high computational cost and slow sampling. More recently, autoregressive generation of handwriting images has also been explored to support text of varying lengths \cite{ref20}. However, these approaches still struggle to capture global, writer-specific structures that define handwriting style, and they lack interpretability tools for understanding how attention mechanisms contribute to style replication. Addressing these limitations requires a unified architecture capable of modeling long-range dependencies across multiple style examples while remaining transparent in its style-content fusion process.

To address these challenges, we propose a series of contributions summarized as follows:
\begin{itemize}
    \item We introduce a ViT-based style encoder to effectively capture global stylistic patterns such as text slant, roundness, and stroke width across multiple style images.
    \item We perform style-content entanglement through cross-attention, which fuses the learned global stylistic patterns from the style images into the target text, enabling synthesis that fully adheres to the provided styles.
    \item We employ an interpretability framework, Salient Stroke Attention Analysis (SSAA), to provide stroke-level insights into how stylistic attributes are transferred during synthesis, offering a clearer understanding of the model’s internal decision-making process.
\end{itemize}

\section{Related Work}

The trajectory of handwriting image generation research generally follows two directions. The first approach captures and processes handwriting as a spatio-temporal sequence of data points rather than a static image. The data stream is a sequence of vectors, where each vector at time step $t$ might contain $(x_t, y_t)$ coordinates, a timestamp and a binary pen-state variable indicating whether the pen is in contact with the surface or lifted \cite{ref8, ref9, ref10, ref11, ref12}. The second approach operates directly on the pixel images of handwritten text and produces bitmap handwriting images \cite{ref3, ref13, ref14, ref15, ref2, ref16}. The former is referred to as the online method, while the latter is known as the offline method.

\subsection{Online Handwriting Generation}
Preliminary online handwriting generation methods employ Recurrent Neural Networks (RNNs) like LSTM and C-VRNN to predict future stroke points sequentially based on previous pen positions and input text \cite{ref8, ref9}. Subsequent work \cite{ref10} focuses on overcoming the sequential generation process and later works \cite{ref11, ref12} introduce the GAN framework with an adversarial learning paradigm for online handwriting generation.

\subsection{Offline Handwriting Generation}
To overcome the difficulty posed by online handwriting generation methods, numerous works have been done to develop offline generation methods which rely solely on pixel-level image data. Several works \cite{ref3, ref5, ref13, ref14, ref15} have been done that utilize CNN-based GAN architecture for handwriting image generation. These approaches follow adversarial setup for training generation models and also utilize recognizer and style classification models to preserve content and style information. However, these approaches fail to effectively capture local writing style and patterns. To overcome this problem, Bhunia et al.\ \cite{ref2} introduce handwriting transformers which focus on style-content entanglement through the use of cross-attention between style vector representation and content text representation. Developing upon this, Pippi et al.\ \cite{ref17} modify the content text representation and instead represent text as a GNU Unifont, basically converting a character into a $16 \times 16$ image, enabling the generation of out-of-charset characters. However, these approaches require 15 style image samples from the writer to effectively carry over the desired style in the generated image. Following the success from transformer and GAN architecture for handwriting generation, several works utilizing diffusion models have also been introduced \cite{ref16, ref19}. More recently, an autoregressive style handwriting image generation method has been introduced which utilizes an autoregressive transformer in the decoder to accommodate the generation of variable length handwriting images \cite{ref20}. However, the image generation process is time-consuming due to its autoregressive nature.

\section{Methodology}
\label{methodology}

\begin{figure}[t]
    \centering
    \includegraphics[width=0.8\linewidth]{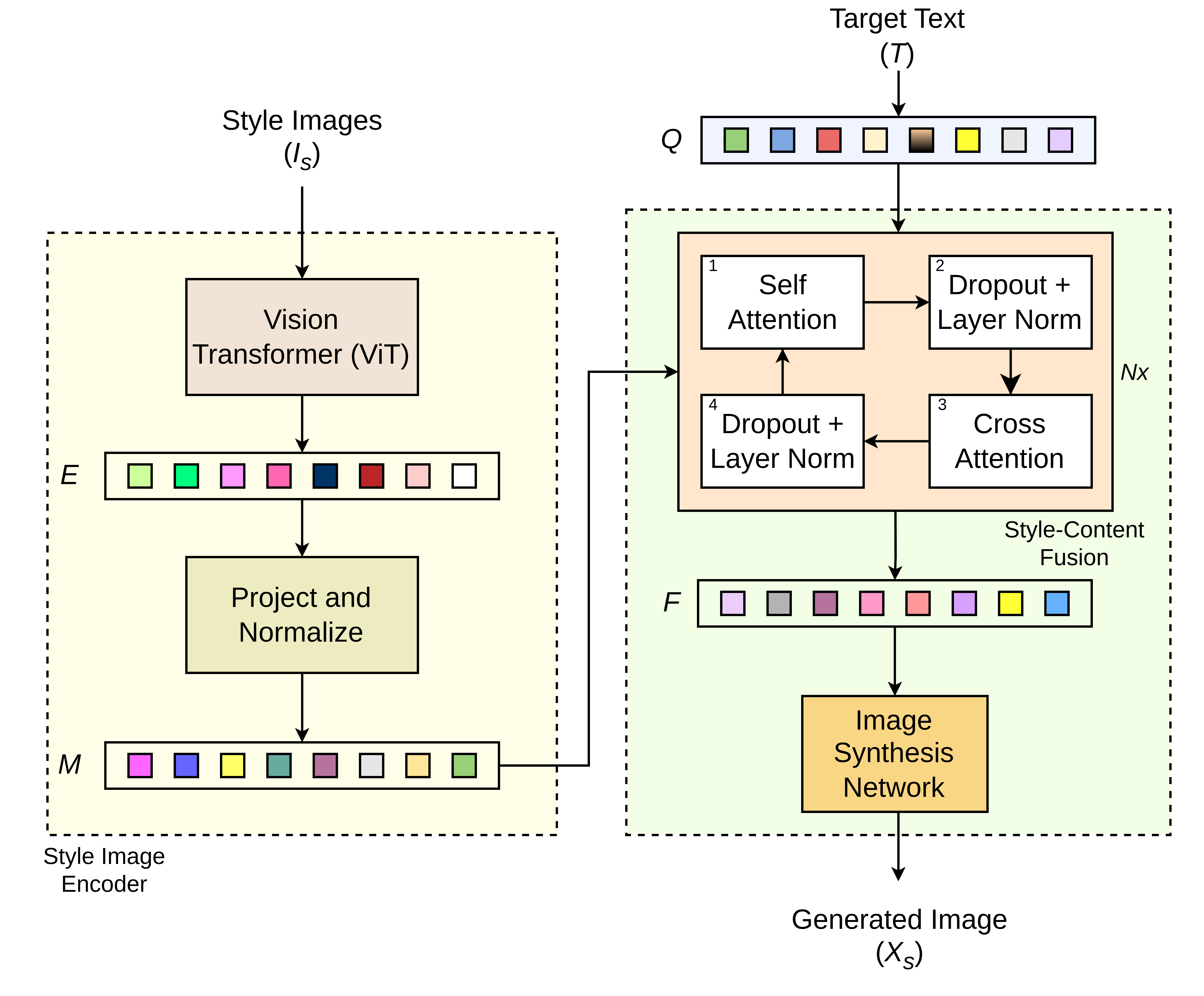}
    \caption{Generator Architecture of ScriptViT}
    \label{fig:block_diagram}
\end{figure}

\subsection{Problem Statement}
\label{problem_statement}
The goal is to synthesize a styled handwriting image $X_s$ containing a user-provided text $T$ using five style samples $I_s$ from a writer $w \in W_s$. The textual content is denoted as $T = \{a_i\}_{i=1}^{L}$, where $L$ is the text length and each $a_i$ belongs to a predefined character set containing alphabets, digits, punctuation marks, and other symbols. Each writer $w$ in the set $W_s$ exhibits a unique handwriting style. The generated image $X_s$ must accurately represent the content text $T$ while replicating the handwriting style associated with writer $w$.

\subsection{Model Architecture}

Our approach to styled handwriting generation is based on a conditional Generative Adversarial Network (cGAN). The system is built on a multi-objective training paradigm first proposed by Bhunia et al.~\cite{ref2}, which is effective at disentangling style and content.

The framework consists of four primary components:

\begin{itemize}
    \item \textbf{Generator $G$:} A conditional network that synthesizes handwritten images based on style and content inputs.
    \item \textbf{Realism Discriminator $D$:} Provides an adversarial objective to ensure that generated images appear realistic, and same as real handwriting.
    \item \textbf{Text Recognizer $TR$:} An OCR network that enforces content fidelity, making sure that the generated writing represents the target text accurately.
    \item \textbf{Writer Classifier Network $WCN$:} A style classifier that enforces style consistency matching target writer’s style with the generated handwriting.
\end{itemize}

While we train this multi-objective training strategy, our main contribution lies in redesigning the Generator’s style encoder. We replace the original hybrid CNN-transformer encoder with a single Vision Transformer \cite{ref25}. By processing style images as a global sequence of patches, ViT is able to capture non-local and calligraphic patterns from the handwriting. 

The generator $G$ is an encoder-decoder network that takes a set of $N=5$ style images and a target text $T$ as input, and synthesizes a handwriting image emulating the given writer’s calligraphic style. It is responsible for creating images that look as realistic as possible. The discriminator $D$ acts as a teacher and guides the generator’s learning process. In addition to the generator and discriminator, we employ a Writer Classifier Network $WCN$ and a Text Recognizer $TR$. The $WCN$ ensures that the generated images carry the desired style, while the $TR$ verifies that they accurately represent the desired text.

\subsubsection{Style Image Encoder}

The Style Image Encoder converts $N=5$ style images $I_s^{B \times C \times H \times W}$ into a single, comprehensive ‘style memory’ bank. Here $B$ represents the batch size and $C$, $H$ and $W$ represent the dimensions of an RGB image. This module uses a Vision Transformer (ViT) as its backbone for representing style images, and divides each style image into a grid of patches. These patches are flattened and linearly projected into embeddings, which are then passed through the encoder blocks of ViT. The self-attention mechanism within the ViT allows the model to compare every patch with every other patch, effectively capturing global stylistic patterns. The ViT produces output embeddings $E^{S \times Q \times 384}$, where $S=B \cdot N$ and $Q$ denotes the number of tokens in the image. We discard the $CLS$ token and only consider the remaining tokens representing patches $P = Q-1$, which are then projected through a linear layer to match the hidden dimension of 512 required by the subsequent module:

\begin{equation}
E_{\text{proj}} = \text{LayerNorm}(\text{Linear}(E[:, 1:, :]))
\end{equation}

Finally, the projected embeddings are normalized and reshaped into the shape of $M^{S_{\text{len}} \times B \times 512}$, where $S_{\text{len}}$ denotes the sequence of patches of $N$ style images concatenated together, i.e., $S_{\text{len}} = N \cdot \text{numpatches}$. This sequence of embeddings serves as a memory bank that contains comprehensive, globally-aware style information of the writer which we call \textbf{Style Memory}.

\subsubsection{Content Query Encoder}

The Content Query Encoder processes the target text string $T$. The text is first tokenized into $K$ characters. Each character token is projected into a latent space of dimension $D_{\text{model}}$ via a learnable embedding layer. The resulting content query $Q^{K \times B \times D_{\text{model}}}$ is obtained.

This sequential representation can be termed as query positional embeddings. They preserve the order of the characters and provide structured input for the decoder. 

\subsubsection{Style-Content Fusion Core}

This is the heart of the generator, where the learned style is applied to the target text. It integrates the Style Memory $M$ with the Content Query $Q$ using a standard Transformer Decoder architecture consisting of $N_x = 3$ layers. The transformer decoder utilizes self-attention and cross-attention in a multi-headed fashion with number of heads $H = 8$. The key operation is cross-attention, where the Content Query $Q$ “attend to” the Style Memory $M$. The style memory $M$ serves as key and value during the cross-attention:

\begin{equation}
\text{Attention}(q, k, v) = \text{softmax}\left(\frac{q k^T}{\sqrt{D_{\text{model}}}}\right)v
\end{equation}

where $q, k,$ and $v$ represent query, key, and value, respectively.

In this process, for each character to be generated, the model examines all style features and selects the most relevant ones. It generates a new sequence of embeddings $F$, where each character embedding is now infused with the appropriate calligraphic style.

\subsubsection{Image Synthesis Network}

This module takes the abstract, style-infused feature vectors and renders them into a final image. It translates the feature sequence from the Fusion Core into a handwritten word image. The sequence is first expanded through a linear layer and reshaped into a 4D tensor suitable for convolutional processing. This tensor is then passed through a series of residual convolutional blocks and upsampling layers that progressively construct the image. This ultimately gives the handwritten image $X_s$ containing the content text $T$ in the style of the desired writer $w$.

\section{Training and Loss Functions}

The model is trained using an adversarial framework following the training process of GAN \cite{ref23}. In this setup, the generator $G$ synthesizes images, while the discriminator $D$ attempts to distinguish real images from the generated ones, thereby pushing the generator to generate more realistic outputs. We use a hinge version of adversarial losses to train both the generator and the discriminator which are defined as follows:

\begin{equation}
L_{\text{G}} = \mathbb{E}[-D(G(I_s, T))]
\label{eq:adv_gen}
\end{equation}

\begin{equation}
L_{\text{D}} = \mathbb{E}_{x \sim I_s}[\max(1 - D(x), 0)] + \mathbb{E}[\max(1 + D(G(I_s, T)), 0)]
\label{eq:adv_dis}
\end{equation}

where, $L_G$ and $L_D$ denote the hinge loss for generator and discriminator, respectively.

Besides realistic image generation, it is also essential to preserve the textual content and the calligraphic styles of the writer. We utilize a Text Recognizer $TR$ network to preserve the textual content in the generated handwriting images. The $TR$ network guides the generator to generate images with textual content that matches the target/content text. We use Connectionist Temporal Classification $CTC$ loss to achieve that objective which is defined as,

\begin{equation}
L_{\text{TR}} = \mathbb{E}_{x \sim \{I_s, X_s\}}[-\sum \log(p(y_r \mid TR(x)))]
\label{eq:tr}
\end{equation}

where, $y_r$ is the transcription of $x \sim \{I_s, X_s\}$

Furthermore, it is also essential to guide the generator in such a way that it is able to clone the writer's style of writing in the generated image. We employ a Writer Classifier Network $WCN$, which provides feedback to the generator by predicting the writer identity from the generated outputs. This encourages the generator to align its outputs with the desired calligraphic style. The $WCN$ is trained as a multi-class classification model with each class representing a writer and optimized using a cross-entropy loss function. It is defined as follows:

\begin{equation}
L_S = \mathbb{E}_{x \sim \{I_s, X_s\}}[-\sum_{i=1}^{W_s} y_i \log(WCN(x))]
\label{eq:ls}
\end{equation}

It is important to note that the $TR$ and $WCN$ models are trained solely on real images. However, the losses computed by these networks on generated images are utilized to guide the training of the generator. The networks are updated in a staggered manner within each epoch:

\begin{enumerate}

\item The Generator $G$ is updated every two iterations utilizing feedback from the Discriminator $D$, $TR$ network and $WCN$, based on the generated images. The total loss is defined as follows:

\begin{equation}
L_{Total} = L_G + L_{\text{TR}}^{\text{fake}} + L_S^{\text{fake}}
\label{eq:gupdate}
\end{equation}

Where, $L_{\text{TR}}^{\text{fake}}$ and $L_S^{\text{fake}}$ denote text recognition loss and writer classifier loss for generated images, respectively.

\item The Discriminator $D$, $TR$ network and $WCN$ are updated every iteration. The discriminator is optimized using adversarial losses derived from both real and synthesized images, whereas the $TR$ network and $WCN$ are trained exclusively on real images. That is,

\begin{equation}
L_{Total} = L_D + L_{\text{TR}}^{\text{real}} + L_S^{\text{real}}
\label{eq:dupdate}
\end{equation}

Where, $L_{\text{TR}}^{\text{real}}$ and $L_S^{\text{real}}$ denote text recognition loss and writer classifier loss for real images, respectively.

\end{enumerate}

\section{Experiments}

We conduct both training and testing on the IAM handwriting dataset \cite{ref4}, which contains handwriting images of various texts in the English language. The dataset consists of a total of 92,369 word images written by 500 different writers, which are further divided into train and test sets. The training set contains 66,603 word images from 339 writers while the test set consists of 25,766 images from 161 different authors. In all the experiments, the images are resized to the fixed height of 32 pixels for training the text recognizer and writer classifier models. However, to meet the input requirements of the Vision Transformer (ViT) model, we transform the style images to the fixed dimension of 224x224 pixels. We do so by padding the style images with white pixels instead of resizing them in order to prevent the style images from getting distorted. In contrast to previous approaches \cite{ref3, ref2}, which required 15 style images as an input, we use only $N=5$ style images in our entire experiment. This reduction in the number of style images was made possible through the use of ViT as the style encoder, which is highly efficient in capturing global stylistic patterns than the traditional CNN models \cite{ref24}. The transformer decoder utilizes 3 attention layers with each multi-headed attention consisting of 8 attention heads. The embedding dimension is 512. We use the Adam optimizer for all networks with $\beta_1 = 0.0$ and $\beta_2 = 0.999$. The learning rates for the generator (G), discriminator (D), writer classifier (WCN), and TR network are all set to $5e^{-5}$. The model is trained for a total of 30 epochs with a batch size of 16 on a CUDA-enabled GPU device.

\section{Evaluation Metrics}

We test the performance of our model across various evaluation metrics to evaluate different aspects of the HTG task. Following the approach used by \cite{ref20}, we use Fréchet Inception Distance (FID) \cite{ref1}, Kernel Inception Distance (KID) \cite{ref7}, Handwriting Distance (HWD) \cite{ref6} and Absolute Character Error Rate Difference ($\Delta$CER) to evaluate our model. FID and KID are metrics commonly used to evaluate the output of generative models in image synthesis. They usually extract features from an intermediate layer of a pre-trained network like InceptionV3 and compute how close the generated image distribution is to a reference distribution. To compute $\Delta$CER, we consider the state-of-the-art text recognizer TrOCR-Base model \cite{ref18}. The $\Delta$CER measures the readability of the generated handwriting images. In all of these metrics, a lower score represents better performance.

\section{Results: Styled Handwriting Generation}

\begin{table}[t]
\centering
\caption{Comparison between our model and previous approaches on the IAM Words dataset across various evaluation metrics. The KID is multiplied by $10^3$ and the best performance is represented in bold.}
\label{tab:results}
\begin{tabular}{lcccc}
\hline
Method & FID$\downarrow$ & KID$\downarrow$ & $\Delta$CER$\downarrow$ & HWD$\downarrow$ \\
\hline
TS-GAN~\cite{ref13} & 129.57 & 141.08 & 0.28 & 4.22 \\
HiGAN+~\cite{ref22} & 50.19 & 43.39 & 0.20 & 3.12 \\
HWT~\cite{ref2} & 27.83 & 19.64 & 0.15 & 2.01 \\
VATr~\cite{ref17} & 30.26 & 22.31 & \textbf{0.00} & 2.19 \\
VATr++~\cite{ref21} & 31.91 & 23.05 & 0.07 & 2.54 \\
One-DM~\cite{ref19} & 27.54 & 21.39 & 0.10 & 2.28 \\
Emuru~\cite{ref20} & 63.61 & 62.34 & 0.19 & 3.03 \\
ScriptViT (Ours) & \textbf{27.02} & \textbf{17.79} & 0.27 & \textbf{1.58} \\
\hline
\end{tabular}
\end{table}

Table~\ref{tab:results} represents evaluation metrics obtained by various approaches for HTG. We utilize the framework provided by HWD~\cite{ref6} to generate these metrics for our approach. The testing is done on a subset of the test set of the IAM dataset, the subset consisting of 17,729 images from 161 different writers. We present the scores provided in~\cite{ref20} for all the previous approaches.


Our model, which employs a ViT-based style feature extractor combined with a Transformer Decoder-based fusion core, surpasses previous state-of-the-art approaches, as shown in Table~\ref{tab:results}. We achieve performance improvements of \textbf{+21.39\%} and \textbf{+9.42\%} in terms of HWD and KID scores, respectively, while attaining comparable performance in terms of FID score as compared to the previous state-of-the-art models that achieved the best results for these metrics.

\subsection{Qualitative Analysis}


\begin{figure}[t!]
    \centering
    \begin{subfigure}{0.48\linewidth}
        \centering
        \includegraphics[width=\linewidth]{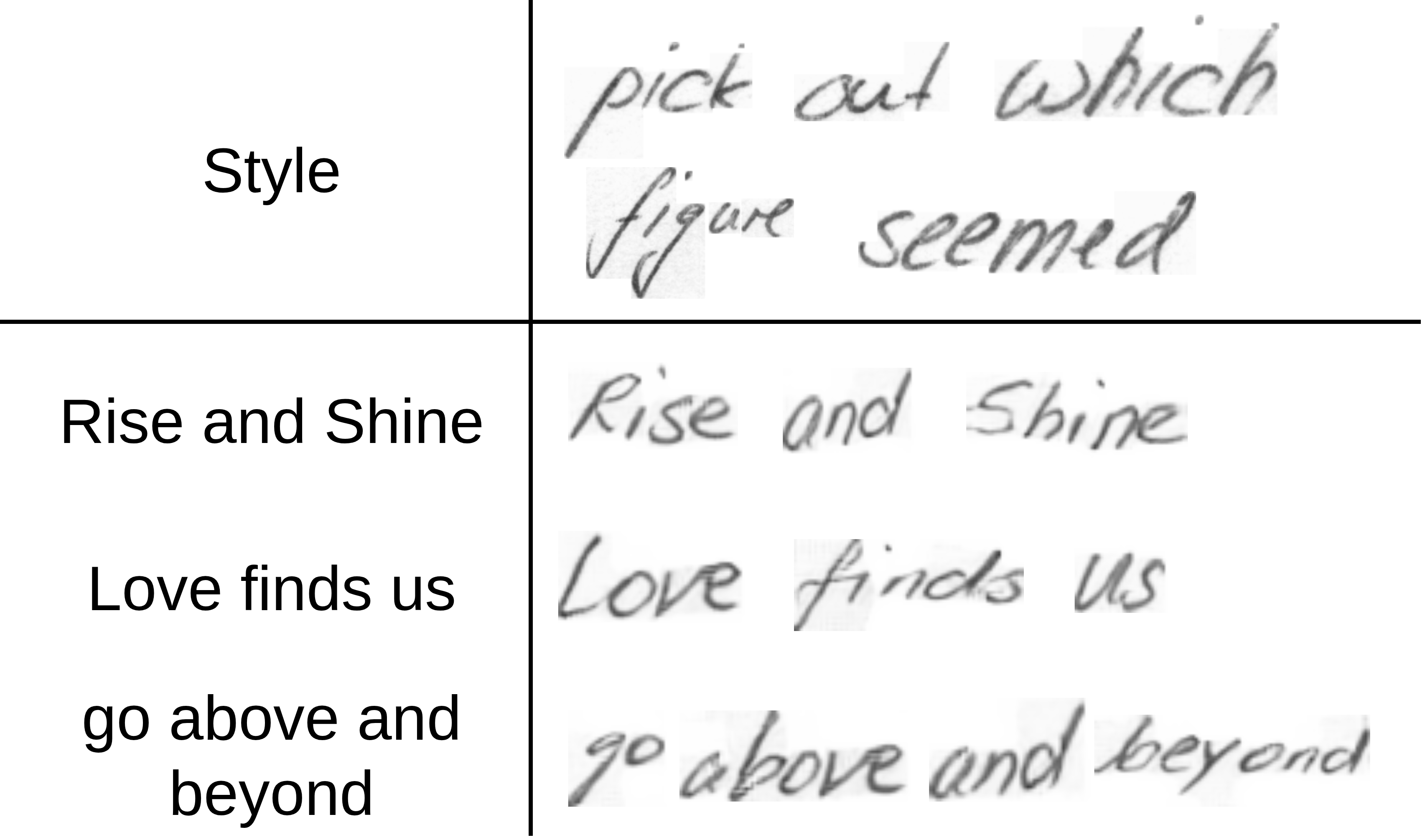}
        \caption{}
        \label{fig:same_writer_diff_text}
    \end{subfigure}
    \hfill
    \begin{subfigure}{0.48\linewidth}
        \centering
        \includegraphics[width=\linewidth]{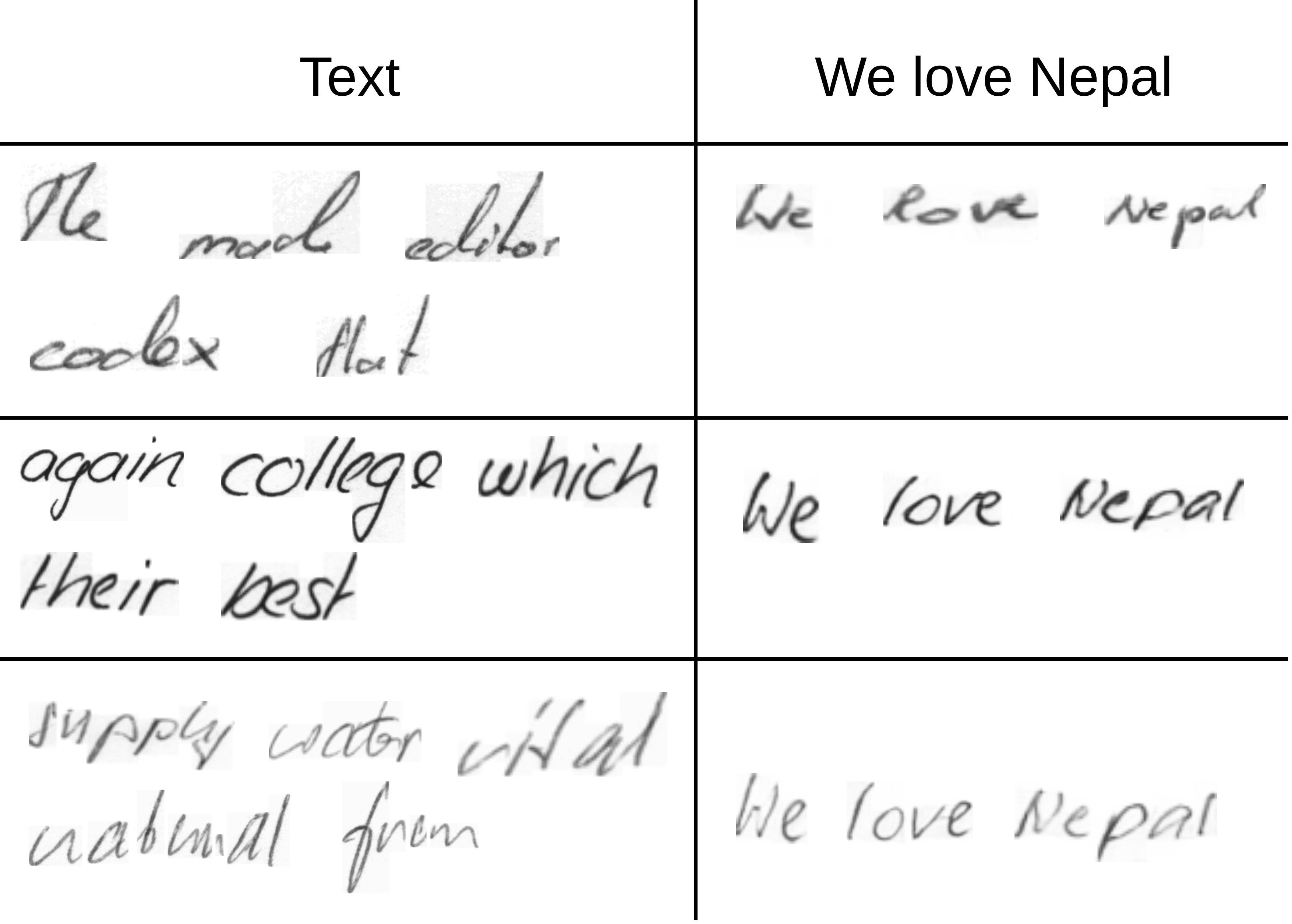}
        \caption{}
        \label{fig:diff_writer_same_text}
    \end{subfigure}
    \caption{Synthesized handwriting images conditioned on same-style and diverse-style inputs. (a) Evaluation on different text when provided with style images from the same writer. The first row represents the style images. (b) Evaluation on the same target text conditioned on style images from multiple writers. The first column represents style images for each writer.}
    \label{fig:style_diverse}
\end{figure}


In addition to quantitative analysis, we also perform qualitative study of the images generated from our approach. We assess our synthesized handwriting images on the basis of style mimicry and correct content representation.

\paragraph{Style Consistency and Transfer:} To assess intra-writer consistency, we keep the writer fixed while varying the target text. The objective is to see if the writer’s style remains consistent even as the text content changes. Figure~\ref{fig:style_diverse}(a) represents multiple synthesized handwriting text images conditioned on style images from the same writer. We can see that the generated handwriting images are able to preserve the stroke thickness, text slant angle and overall calligraphic style of the provided style images, irrespective of the target text. Moreover, they are also able to replicate the local stylistic patterns present in the style images. In addition to evaluating intra-writer consistency, we also assess our model’s ability to clone diverse styles from different authors. We synthesize the same text, “We love Nepal”, conditioned on a variety of styles and present the result in Figure~\ref{fig:style_diverse}(b). We can see that the generated images completely align with the calligraphic styles of the provided style images. The stroke, boldness and character size present in the style images are all preserved in the generated handwriting images. This demonstrates the robustness of our model in understanding and replicating any custom handwriting style. In addition to style cloning, the generated handwriting images also accurately represent the target content in both cases.

\begin{figure}[t]
\centering
\includegraphics[width=\linewidth]{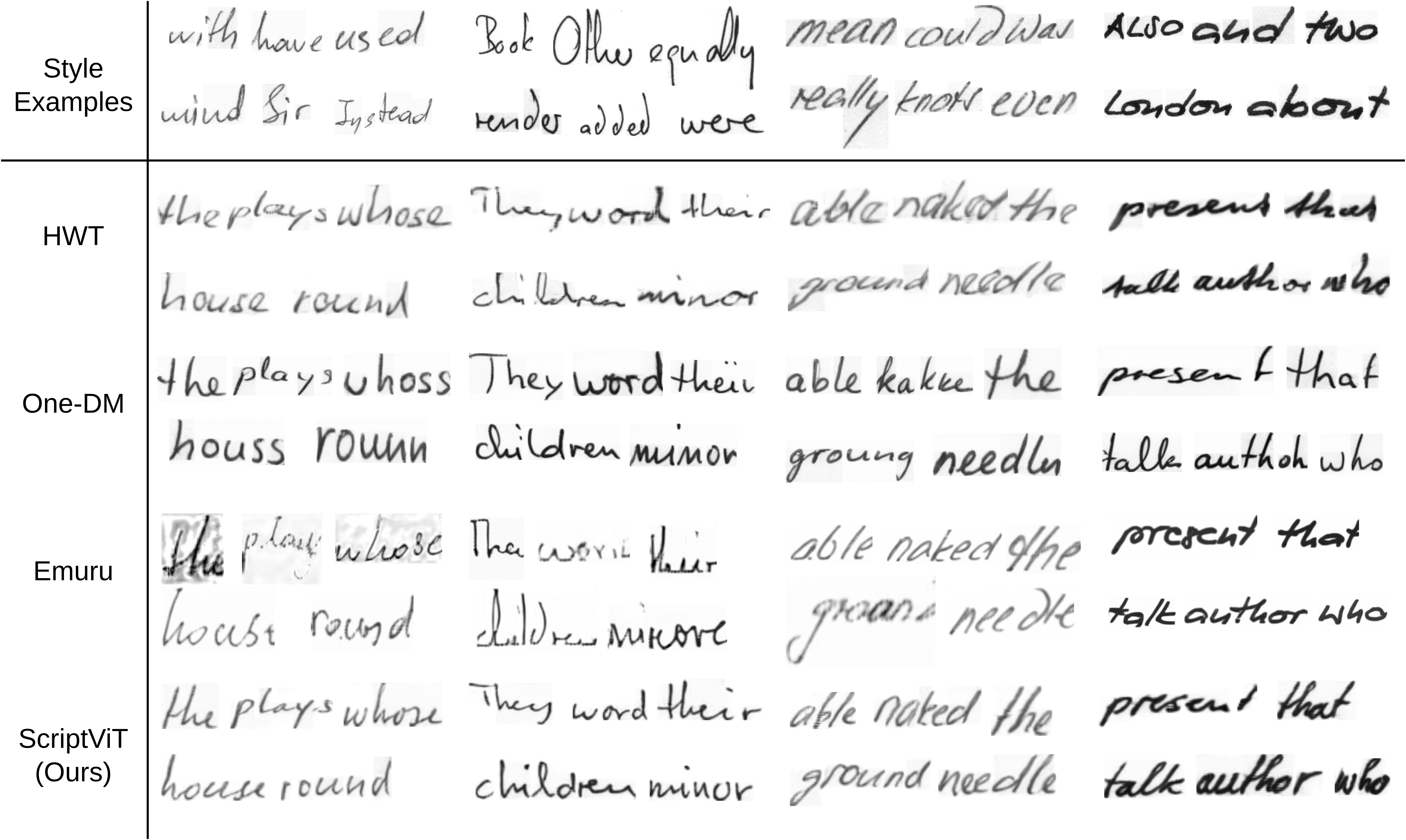}
\caption{Qualitative comparison between our model and previous approaches on Handwritten Text Generation.}
\label{fig:qual_comparison}
\end{figure}

We also present a comparison between the handwriting images generated from our approach and those produced by previous state-of-the-art methods in Figure~\ref{fig:qual_comparison}. We can see that our approach is able to replicate exact stroke thickness, character slants and overall style patterns, validating its ability to capture diverse calligraphic styles. In contrast, previous approaches like One-DM and Emuru seem to struggle with accurately capturing given style patterns. One-DM has difficulty capturing global style patterns, resulting in incorrect identification of stroke thickness. Emuru appears to struggle with correctly rendering the target text and also produces noisy images. HWT produces satisfactory images in terms of content accuracy and local calligraphic features but struggles to capture global stylistic patterns.

\subsection{Interpretability: Salient Stroke Analysis}

Understanding how the model performs the calligraphic style transfer requires identifying which specific handwriting features contribute most to the generated output.

To address this, we implement Salient Stroke Attention Analysis (SSAA). It is a visualization method that projects the decoder's attention weights back onto the ink regions of the style images. This makes it possible to identify the specific stroke patterns that contribute to stylistic adaptation and to interpret how the model integrates these features during synthesis. The process has five separate stages:

\paragraph{Word-Level Attention Extraction:} For each generated word, we extract the cross-attention weights from the final layer of the Style-Content Fusion Core. 
\begin{equation}
A \in \mathbb{R}^{H \times K \times L}
\end{equation}
where
\begin{itemize}
    \item $H$ is the number of attention heads,
    \item $K$ is the number of query tokens (characters) in the generated word,
    \item $L$ is the number of key tokens corresponding to patches in the reference style images.
\end{itemize}

To obtain a unified representation of which regions influence the generated handwriting, we average the attention weights across both heads and character positions:

\begin{equation}
a_{\text{word}}[l] = \frac{1}{H \cdot K} \sum_{h=1}^{H} \sum_{k=1}^{K} A[h,k,l], \quad l = 1, \ldots, L
\end{equation}

This gives a single attention vector $a_{\text{word}}$ assigning a single importance score to each patch, reflecting its contribution to the formation of the target word.

\paragraph{Attention Map Reconstruction:} This attention vector is divided into parts corresponding to each style example. Each part is reshaped into a patch grid and upsampled to match the original image resolution, resulting in a continuous attention map. The maps are then normalized to the range $(0,1)$ for consistency.

\paragraph{Masked Attention on Ink (MAI):} Since background areas do not contribute to style transfer, we isolate only the ink regions. Each style image is converted to grayscale, binarized using Otsu’s thresholding, and cleaned with a median filter. The resulting mask is applied to the attention map so that only the inked strokes are retained.

\paragraph{Salient Stroke Identification:} To highlight the most influential regions, we select attention values above a chosen percentile threshold and apply connected-component analysis to group neighboring pixels into strokes. Small regions are removed, and the largest remaining components are treated as the most salient strokes contributing to the generated output.

\paragraph{Visualization Grid:} For interpretability, we overlay a semi-transparent highlight on the style images. This visualization directly links decoder attention to the inked regions, allowing precise interpretation of which strokes and features are transferred during the process of handwriting generation.

\begin{figure}[t]
    \centering
    \includegraphics[width=0.8\linewidth]{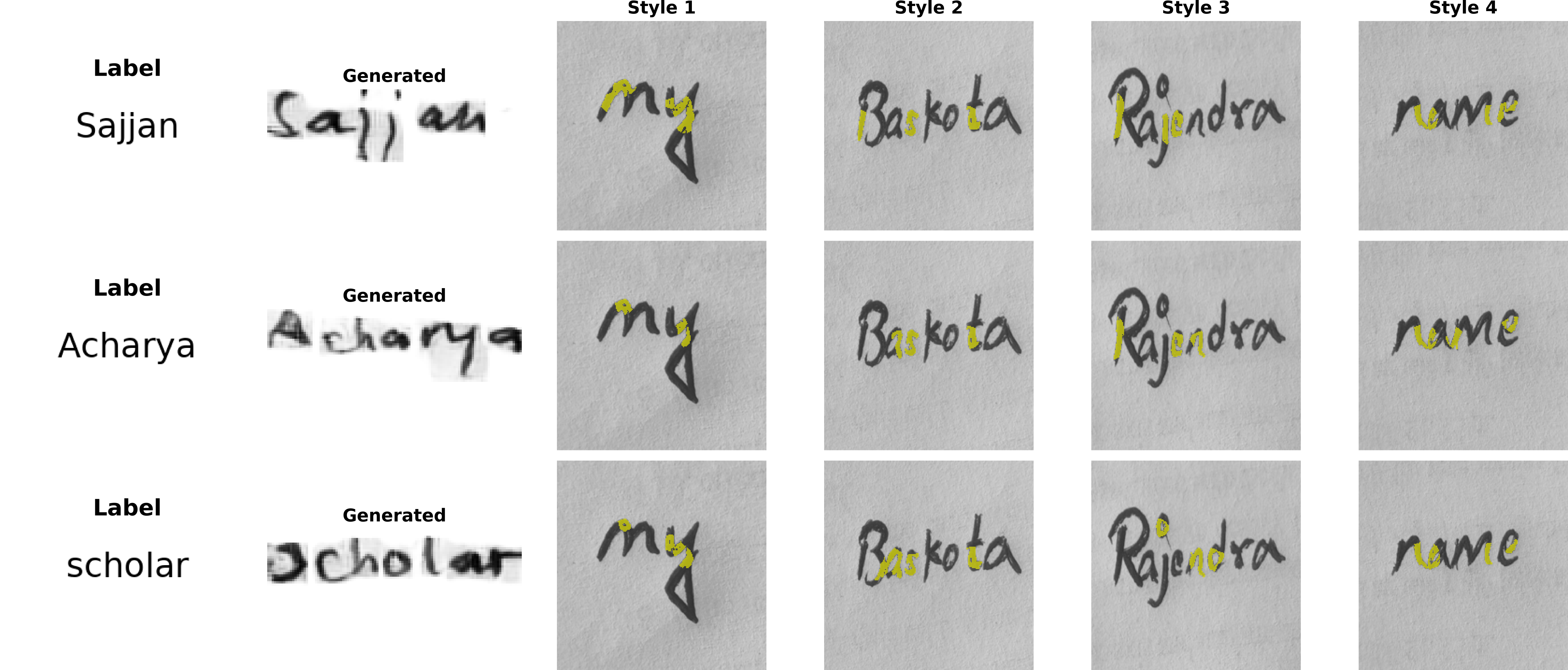}
    \caption{Attention Visualization on Style Images for each generated word.}
    \label{fig:attention_vis}
\end{figure}

From Figure~\ref{fig:attention_vis}, it can be stated that the model is not performing a 1-to-1 letter copy. Instead, it decomposes the target word ‘scholar’ into its abstract components:

\begin{itemize}
    \item An “initial capital stroke”
    \item “Tall vertical stems” (for ‘h’ and ‘l’)
    \item “Rounded and connected strokes” from the m/n/a letters from the style images
\end{itemize}

The generator intelligently identifies the ascenders, and recognizes rhythmic strokes, leading to a stable and consistent generated style. The ViT encoder creates a rich, patch-based vocabulary of style features, which are then queried by the transformer's decoder's cross-attention, and are applied as per the requirement.

\section{Conclusion}

In this work, we presented ScriptViT, a unified framework for personalized offline handwriting generation that combines a ViT-based style encoder, a transformer decoder for style-content fusion and an interpretability module utilizing Salient Stroke Attention Analysis. Our model is able to effectively capture both local stroke details and global calligraphic patterns such as slant, curvature, stroke thickness etc. It does so by encoding multiple style images into a global style memory and injecting these features into the target text through a cross-attention mechanism. Experiments on the IAM words dataset reveal that ScriptViT achieves performance improvements across standard HTG metrics, including HWD and KID scores and comparable FID relative to prior transformer and diffusion based approaches.

\bibliographystyle{unsrtnat}
\bibliography{references}







\end{document}